\documentclass[conference]{IEEEtran}
\usepackage{cite}
\usepackage{amsmath,amssymb,amsfonts}
\usepackage{graphicx}
\usepackage{xcolor}
\usepackage{booktabs}
\usepackage{multirow}
\usepackage{array}
\usepackage{etoolbox}
\usepackage{acronym}

\usepackage[final]{hyperref}
\hypersetup{
 colorlinks=true,
 linkcolor=blue,
 filecolor=magenta,
 urlcolor=blue,
 citecolor=black
}

\acrodef{DOR}{Dynamic Object Removal}
\providecommand{\figref}[1]{Fig.~\ref{#1}}
\providecommand{\tabref}[1]{Table.~\ref{#1}}
\apptocmd{\thebibliography}{\small}{}{}

\begin{document}

\title{ELite++: Efficient Lifelong LiDAR Mapping via Window-Based Registration and Voxel Ratio Maps}

\author{\IEEEauthorblockN{Geonhui Lee${}^{1}$, Hyeonjae Gil${}^{1}$, Ayoung Kim${}^{1}$ and Minwoo Jung${}^{1*}$ }
\IEEEauthorblockA{\textit{${}^{1}$Department of Mechanical Engineering, Seoul National University},\\
Seoul, 08826, Korea ([geongeon, h.gil, ayoungk, moonshot]@snu.ac.kr)}
}

\maketitle

\begin{abstract}
Long-term LiDAR mapping requires clean and consistent static maps across repeated sessions in dynamic environments.
Existing lifelong mapping pipelines combine session alignment, dynamic filtering, and map update, but repeated dense point-wise updates and candidate-dependent alignment can limit scalability.
To tackle these issues, we present ELite++, an efficient extension of ELite that combines automatic window-based registration, aligned-session dynamic object removal, and compact Voxel Ratio Map integration.
ELite++ removes transient objects while preserving static structures through hit-exposure voxel stability, local ground protection, and structure-aware refinement.
Experiments on diverse LiDAR datasets show improved dynamic object removal over existing DOR methods, preserved multi-session alignment quality, and runtime reduced to 11\% of the original pipeline.
\end{abstract}


\section{Introduction}

Accurate localization and consistent map maintenance are fundamental for autonomous navigation, and long-term operation further requires maps to adapt to environmental changes over time~\cite{cadena2016past,sousa2023systematic}.
In real-world LiDAR mapping, repeated sessions often contain odometry drift, partial overlap, and scene changes from moving or temporarily parked objects; directly accumulating them leaves transient artifacts that degrade localization and spatial analysis.
Lifelong mapping addresses this by aligning new sessions to an existing map and maintaining persistent static structures over time~\cite{kim2021ltmapper,gil2025elite}.

Achieving this goal requires not only aligning sessions, but also deciding which observations should remain in the map.
Although multi-session SLAM and map-merging systems improve alignment across sessions~\cite{huang2021disco,yin2023automerge,lim2025multi,wei2024large}, they do not explicitly maintain a clean static representation under dynamic scene changes.
ELite~\cite{gil2025elite} addresses this issue through local and global ephemerality, enabling short- and long-term transiency modeling.
However, it still relies on reliable initial session-to-map associations, and repeated dense point-wise ephemerality updates can become costly as sessions accumulate.
This motivates an efficient formulation that automatically verifies session alignments and accumulates long-term static evidence as compact voxel-level statistics.

Before such static evidence is committed to the maintained map, transient observations must be reliably filtered.
Existing \ac{DOR} methods remove many dynamic traces by exploiting temporal consistency and visibility cues~\cite{hornung2013octomap,schauer2018peopleremover,kim2020remove,lim2021erasor,lim2023erasor2,fan2022dynamicfilter}.
However, local visibility or voxel-wise decisions may still cause ground erosion, bottom residues, and fragmented misclassifications, which can accumulate as persistent artifacts in lifelong mapping.
Thus, \ac{DOR} should operate on aligned multi-view observations and support long-term static evidence accumulation across sessions.

To address these challenges, we propose ELite++, an efficient lifelong mapping framework for static map maintenance.
ELite++ combines automatic window-based session registration, aligned-session \ac{DOR}, and compact Voxel Ratio Map integration.
The registration module verifies local query submaps, fuses trusted window-level poses, and interpolates low-confidence intervals, while the \ac{DOR} module estimates hit-exposure voxel stability with ground-aware and point-level refinements.
Instead of repeatedly updating dense point-wise ephemerality, Voxel Ratio Maps accumulate voxel-level hit and exposure statistics across sessions, reducing update time while preserving alignment quality and static map cleanliness.
The main contributions of this paper are summarized as follows:
\begin{itemize}

    \item We present ELite++, an efficient extension of ELite that automates window-based registration and maintains compact Voxel Ratio Maps, reducing alignment effort and enabling faster lifelong updates.

    \item We propose a ground-aware \ac{DOR} module that reduces ground erosion, bottom residues, and static-structure loss, providing cleaner voxel evidence for robust map integration.

    \item We validate ELite++ on diverse LiDAR datasets, showing improved DOR performance, preserved alignment quality, and runtime reduced to 11\% of the original pipeline for continuous lifelong mapping.
\end{itemize}

\section{Related Works}
\subsection{Multi-Session Map Maintenance}

Aligning point clouds from multiple robots or temporal sessions is commonly addressed as a multi-session SLAM or map-merging problem.
Loop-closure-based methods, such as DiSCo-SLAM~\cite{huang2021disco} and AutoMerge~\cite{yin2023automerge}, detect inter-session overlaps and optimize the resulting constraints through pose graph optimization or global smoothing.
Recent frameworks such as Multi-Mapcher~\cite{lim2025multi} further reduce reliance on explicit loop closures using large-scale map-to-map registration and anchor node-based optimization.
While these methods achieve effective inter-session alignment and globally consistent map construction, their main focus remains geometric consistency rather than static map maintenance under dynamic scene changes.

Lifelong mapping methods further consider repeated map updates in changing environments.
LT-mapper~\cite{kim2021ltmapper} and ELite~\cite{gil2025elite} address LiDAR-based lifelong mapping through modular map maintenance and ephemerality-based map updating, respectively, while LAMM~\cite{wei2024large} incorporates \ac{DOR} into multi-session point-cloud map fusion.
These frameworks show the importance of combining alignment and map cleaning, but their multi-stage processing can increase update time in large-scale or repeated-session settings.
Moreover, maintaining a clean static map remains challenging when dynamic filtering suffers from visibility ambiguity, ground geometry, or residual artifacts near dynamic objects.

Building on this perspective, our framework treats multi-session registration as the first stage of static map maintenance rather than as an independent alignment objective.
The proposed window-based registration aligns local query submaps to the maintained map, and the aligned session is subsequently used for spatio-temporal \ac{DOR}.
This coupling allows transient objects to be filtered before map integration, supporting cleaner lifelong map updates.

\subsection{Dynamic Object Removal}

\ac{DOR} is essential for constructing static LiDAR maps, as moving vehicles, pedestrians, and temporarily observed objects can remain as ghost artifacts.
Early occupancy- and visibility-based approaches discretize space and infer temporal consistency from sensor observations.
OctoMap~\cite{hornung2013octomap} introduced probabilistic octree-based occupancy mapping, while Peopleremover~\cite{schauer2018peopleremover} removes dynamic objects through voxel traversal.
Range-image and visibility-based methods, such as Removert~\cite{kim2020remove}, further construct static maps by removing dynamic traces and recovering falsely removed static points.

Recent methods improve static mapping with ground reasoning, pseudo-occupancy, instance-level reasoning, and online map consistency.
ERASOR~\cite{lim2021erasor} identifies dynamic regions using egocentric pseudo-occupancy and ground-aware analysis, while ERASOR2~\cite{lim2023erasor2} extends this direction with instance-aware dynamic point rejection.
Other approaches combine ground segmentation with occupancy mapping~\cite{arora2021mapping} or perform online dynamic removal through scan-to-map and map-to-map reasoning~\cite{fan2022dynamicfilter}.
Despite these advances, visibility- or voxel-based decisions can still suffer from ground erosion, bottom residues, and isolated misclassifications.

Accordingly, our \ac{DOR} module addresses these failure cases through spatio-temporal voxel analysis on the aligned session.
It combines repeated hit-exposure observations, normal-based ground protection, and cluster-level structural refinement to suppress transient objects while preserving persistent static structures before map integration.

\section{Method}
\label{sec:method}

\subsection{System Overview}
\label{subsec:system_overview}

\begin{figure}[t]
    \centering
    \includegraphics[width=\linewidth]{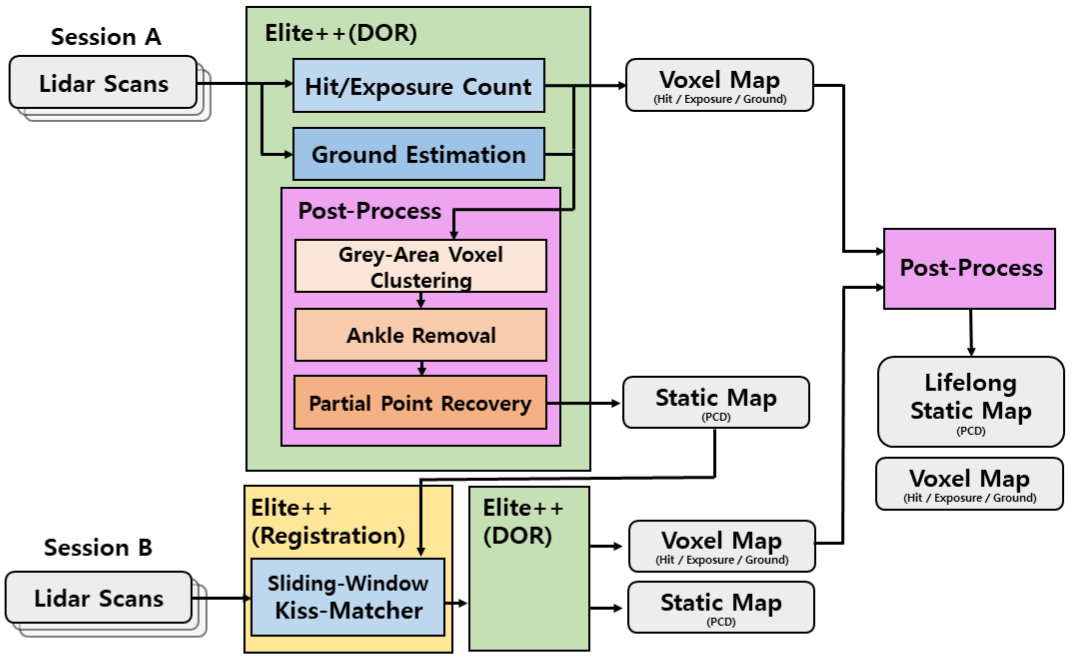}
    \caption{The overview pipeline of ELite++}
    \label{fig:pipeline}
    \vspace{-4mm}
\end{figure}

The proposed lifelong mapping system updates an existing map using newly acquired LiDAR sessions.
Following the general formulation of ELite~\cite{gil2025elite}, the system incrementally maintains a static map over repeated sessions.
We focus on two components that directly affect long-term map quality:
(i) window-based multi-session registration and
(ii) \ac{DOR} before map integration. \figref{fig:pipeline} shows the overview pipeline of ELite++.

Let the $t$-th session be $\mathcal{S}_t=\left\{\left(\mathcal{P}_{t,i}, \mathbf{T}_{t,i}\right)\right\}_{i=1}^{N_t}$, where $\mathcal{P}_{t,i}$ is the $i$-th LiDAR scan and $\mathbf{T}_{t,i} \in \mathrm{SE}(3)$ is its local pose.
The registration module estimates aligned poses $\widetilde{\mathbf{T}}_{t,i}$ and constructs the aligned session map
$\mathcal{M}_{S_t}=\bigcup_i \widetilde{\mathbf{T}}_{t,i}\mathcal{P}_{t,i}$.
The \ac{DOR} module removes dynamic regions from $\mathcal{M}_{S_t}$ and outputs the cleaned static session map $\widehat{\mathcal{M}}_{S_t,\mathrm{st}}$.
The cleaned static session map and its voxel statistics form a session-level Voxel Ratio Map $\mathbf{V}_{S_t}$, which is integrated into the maintained voxel-statistic map $\mathbf{V}_t$.

\subsection{Window-based Multi-session Registration}
\label{subsec:registration}

The registration module estimates a reliable session-to-map transformation under partial overlap, odometry drift, and dynamic scene changes.
Since individual scans provide limited spatial context, we construct overlapping local query submaps and sequentially track their alignments to the existing map.
Trusted window-level alignments are fused to obtain scan-wise poses, while low-confidence intervals are recovered through trajectory interpolation.

The previous map is voxel-downsampled to form a compact registration target $\mathcal{C}_{t-1}$.
For the query session, overlapping temporal windows are defined as $\mathcal{W}_u=\{u,\ldots,u+L-1\}$, where $L$ is the window length and the starting index $u$ is sampled every $r$ scans.
The scans in each window are transformed into the coordinate frame of the first scan and aggregated as
\begin{equation}
    \mathcal{Q}_u
    =
    \bigcup_{i \in \mathcal{W}_u}
    \mathbf{T}_{t,u}^{-1}
    \mathbf{T}_{t,i}
    \mathcal{P}_{t,i}.
\end{equation}
This aggregation provides denser geometric constraints than a single scan and reduces boundary effects through window overlap.

Each query submap $\mathcal{Q}_u$ is sequentially matched against $\mathcal{C}_{t-1}$ using KISS-Matcher~\cite{lim2025kissmatcher}, while the current tracking prior $\mathbf{T}_{\mathrm{init}}$ is updated from trusted windows and carried to the next window.
KISS-Matcher estimates a candidate transformation $\mathbf{T}^{u}_{\mathcal{M}\leftarrow\mathcal{Q}}$ through feature-based global registration and robust outlier pruning.
Since $\mathcal{Q}_u$ is represented in the local frame of scan $u$, the corresponding session-level candidate is $\mathbf{T}^{u}_{\mathrm{abs}}=\mathbf{T}^{u}_{\mathcal{M}\leftarrow\mathcal{Q}}\mathbf{T}_{t,u}^{-1}$.
The candidate is verified by a fitness score $f_u$, the ratio of transformed query points within $\tau_d$ of the target map.
If $f_u \ge \tau_f$, the window is trusted and $\mathbf{T}_{\mathrm{init}}$ is updated by $\mathbf{T}^{u}_{\mathrm{abs}}$.

For each scan $\mathcal{P}_{t,i}$, we collect trusted window estimates covering the scan:
\begin{equation}
    \mathcal{A}_i =
    \left\{
    \mathbf{T}^{u}_{\mathrm{abs}}\mathbf{T}_{t,i}
    \mid
    i \in \mathcal{W}_u,\;
    f_u \ge \tau_f
    \right\}.
\end{equation}
If $\mathcal{A}_i$ is non-empty, the final pose $\widetilde{\mathbf{T}}_{t,i}$ is obtained by fusing the trusted estimates.
Otherwise, it is recovered by distance-weighted interpolation from neighboring trusted poses.
This scan-level fusion avoids reliance on a single selected window and maintains a continuous aligned trajectory across low-confidence intervals.
The resulting aligned session is then passed to the \ac{DOR} module for spatio-temporal analysis in a common coordinate frame.

\subsection{Dynamic Object Removal}
\label{subsec:dynamic_removal}

After registration, the aligned session may still contain dynamic or temporarily observed objects.
We analyze it using a spatio-temporal voxel representation, where candidates are estimated from observation consistency and refined by ground protection, DBSCAN clustering, ankle verification, and intra-voxel restoration.

For each scan index $i$, neighboring aligned scans within a temporal radius $k$ are aggregated into
\begin{equation}
    \mathcal{Z}_{t,i}
    =
    \bigcup_{j=i-k}^{i+k}
    \widetilde{\mathbf{T}}_{t,j}
    \mathcal{P}_{t,j}.
\end{equation}
The submap is discretized into voxels.
For each voxel $v$, we accumulate a hit count $H(v)$ and an exposure count $E(v)$.
The hit count measures direct point observations, while the exposure count is computed by raycasting along the sensor line-of-sight; voxels traversed before a surface return increase their exposure values.
The voxel stability score is
\begin{equation}
    R(v)
    =
    \frac{H(v)}
    {H(v)+E(v)+\eta},
\end{equation}
where $\eta$ is a small constant.
A high $R(v)$ indicates stable occupancy, whereas a low $R(v)$ indicates an unstable dynamic candidate.

\begin{figure*}[t!]
    \centering
    \includegraphics[width=0.8\linewidth]{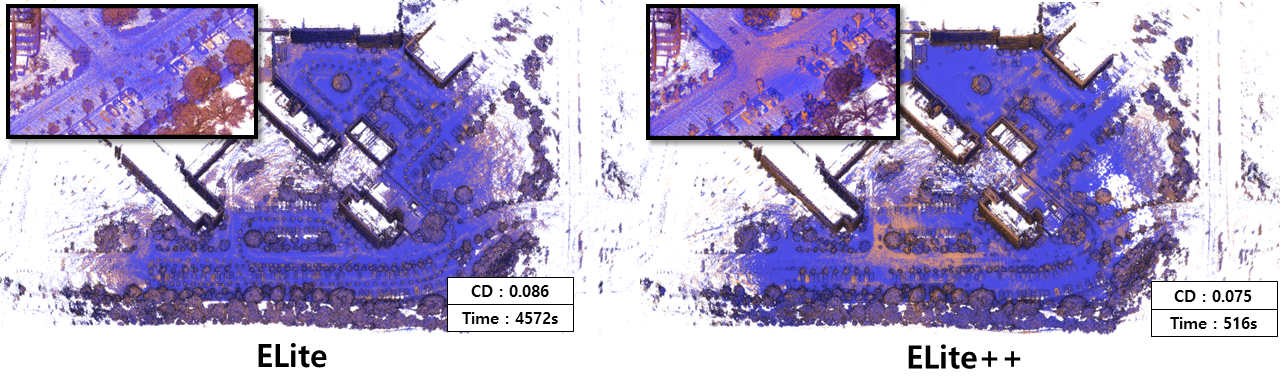}
    \caption{Multi-session static map construction on \texttt{ParkingLot01-02}. ELite++ maintains Chamfer-distance-based alignment quality comparable to ELite while reducing runtime. Zoomed regions show that ELite retains dynamic-object ghost artifacts, whereas ELite++ produces a cleaner static map.
}
    \label{fig:multi}
\end{figure*}

Before dynamic classification, ground regions are protected.
Ground anchors are extracted from points whose normals align with the vertical direction, \textit{i.e.}, $|n_z|>\tau_{\mathrm{slope}}$.
The representative ground height $h_g$ is estimated from these anchors, and voxels satisfying $|v_z-h_g|<\delta_{\mathrm{ground}}$ are excluded from dynamic removal.
This prevents visibility-based filtering from eroding road or terrain surfaces in sloped or sparsely observed regions.
After excluding protected ground voxels, voxels with low stability scores are selected as dynamic candidates.

The remaining candidate voxels are grouped into spatial objects using DBSCAN on voxel centroids with search radius $\epsilon$ and minimum neighborhood size $N_{\mathrm{min}}$.
For each cluster $\mathcal{C}_k$, the average stability score is computed as
\begin{equation}
    \bar{R}(\mathcal{C}_k)
    =
    \frac{1}{|\mathcal{C}_k|}
    \sum_{v \in \mathcal{C}_k}
    R(v).
\end{equation}
Clusters with $\bar{R}(\mathcal{C}_k)<\tau_{\mathrm{cluster}}$ are labeled as dynamic.
This object-level decision suppresses scattered noise and improves transient object removal over voxel-wise classification.

Finally, point-level refinement is applied to handle bottom residues and over-removed static structures.
First, an ankle verification rule examines each protected ground voxel and its vertical upper neighbor $v_{\mathrm{above}}$.
If $v_{\mathrm{above}}$ is dynamic, points inside the ground voxel are re-evaluated by height; points satisfying $p_z>\bar{h}_v+\delta_{\mathrm{tol}}$ are removed as lower-extremity dynamic residues, while the remaining ground points are preserved.
If $v_{\mathrm{above}}$ is static or empty, all points in the ground voxel are locked as static.

Second, intra-voxel point restoration recovers sparse static structures enclosed by dynamic labels.
Isolated dynamic voxels are identified using a 26-neighborhood adjacency search, and their point covariance is analyzed by PCA.
The planarity score is defined as $\Pi_v = (\lambda_2-\lambda_3) / {\lambda_1}$, where $\lambda_1 \ge \lambda_2 \ge \lambda_3 \ge 0$ are the sorted eigenvalues.
Each point $p$ in an isolated dynamic voxel is restored with probability
\begin{equation}
P_{\mathrm{rec}}(p)
=
\min
\left(
1,\,
\frac{R(v)}{2}(1+\Pi_v)
\right).
\end{equation}
This favors recovery of stable and planar structures such as walls or sparse static surfaces, while retaining the voxel-level dynamic label for downstream processing.
After refinement, points in verified dynamic regions are removed from the session map:
\begin{equation}
    \widehat{\mathcal{M}}_{S_t,\mathrm{st}}
    =
    \mathcal{M}_{S_t}
    \setminus
    \widehat{\mathcal{M}}_{S_t,\mathrm{dyn}}.
\end{equation}

\subsection{Static Map Integration}
\label{subsec:map_integration}

The cleaned static representation is integrated using compact Voxel Ratio Maps rather than raw point-cloud fusion.
Similar to the long-term evidence accumulation in ELite~\cite{gil2025elite}, our update maintains persistent observations across sessions.
However, instead of point-wise global ephemerality, we store voxel-level hit and exposure statistics that represent occupancy stability.
For each session $\mathcal{S}_t$, the system stores voxel keys, hit counts $H(v)$, exposure counts $E(v)$, and ground flags.
Let $\mathbf{V}_{t-1}$ and $\mathbf{V}_{S_t}$ denote the historical and current session voxel ratio maps, respectively.

For overlapping voxel keys, a density-adaptive decay factor balances historical evidence with the current observation.
Given $N_{S_t}(v)=H_{S_t}(v)+E_{S_t}(v)$, the historical statistics are scaled by
\begin{equation}
\alpha_v =
\max
\left(
\alpha_{\mathrm{min}},
\min
\left(
\alpha_{\mathrm{max}},
\exp
\left(
-\gamma
\frac{N_{S_t}(v)}{N_{\mathrm{max}}}
\right)
\right)
\right).
\end{equation}
where $N_{\mathrm{max}}$ is the largest session intensity, $\gamma$ controls decay sensitivity, and $\alpha_{\mathrm{min}},\alpha_{\mathrm{max}}$ bound the update range.
The voxel statistics are updated by
\begin{equation}
\begin{aligned}
H_t(v) &= \alpha_v H_{t-1}(v) + H_{S_t}(v), \\
E_t(v) &= \alpha_v E_{t-1}(v) + E_{S_t}(v).
\end{aligned}
\end{equation}
The ground flag is updated from the current session to reflect the latest local terrain observation.

Because integration is performed in voxel-statistic space, the system avoids repeated global nearest-neighbor association or raw point-cloud fusion.
The updated map $\mathbf{V}_t$ is then refined using the same voxel-level filtering rules, allowing dynamic regions and residual noise to be removed before generating the final static map.

\section{Experiments}
\label{sec:experiments}
We evaluate the proposed framework on multi-session alignment and \ac{DOR} using diverse LiDAR datasets: KITTI
(\texttt{KITTI00}, \texttt{KITTI05})~\cite{geiger2013vision}, Argoverse~2
(\texttt{AV2.0})~\cite{Argoverse2}, LTMapper
(\texttt{Parkinglot01-02})~\cite{kim2021ltmapper}, MulRan
(\texttt{KAIST01})~\cite{kim2020mulran}, HeLiPR
(\texttt{KAIST04})~\cite{jung2024helipr}, and CU-Multi
(\texttt{Kittredge Loop}, \texttt{Main Campus})~\cite{albin2025cumulti}.
They cover urban driving, campus-scale mapping, parking-lot environments, and multi-session or multi-robot scenarios.

For multi-session alignment, we report registration accuracy using RMSE and Chamfer Distance (CD).
For \ac{DOR}, we compare against ERASOR~\cite{lim2021erasor}, Removert~\cite{kim2020remove}, OctoMap~\cite{hornung2013octomap}, and OctoMap with ground filtering, using dynamic accuracy (DA), static accuracy (SA), average accuracy (AA), and elapsed time.
All experiments are conducted on a desktop with an AMD Ryzen 5 7500F CPU, GeForce RTX 4060 GPU and 32GB RAM.

\begin{table}[t]
\centering
\caption{Registration Results in Multi-Session Environment}
\label{tab:registration_results}
\footnotesize
\setlength{\tabcolsep}{5.0pt}
\renewcommand{\arraystretch}{1.1}
\resizebox{\columnwidth}{!}{
\begin{tabular}{llccc}
\specialrule{0.8pt}{0pt}{2pt}
Sequence & Method & \multicolumn{3}{c}{Metrics} \\
\cline{3-5}
 & & RMSE $\downarrow$ & CD $\downarrow$ & Time(s) $\downarrow$ \\
\specialrule{0.5pt}{2pt}{2pt}
\texttt{LT-ParkingLot(01-02)} & ELite & \textbf{0.076} & 0.086 & 4572 \\
 & ELite++ (Ours) & 0.084 & \textbf{0.075} & \textbf{516} \\
\specialrule{0.5pt}{2pt}{2pt}
\texttt{KAIST(01-04)} & ELite & 0.106 & \textbf{0.446} & 12732 \\
 & ELite++ (Ours) & \textbf{0.103} & 0.541 & \textbf{1205} \\
\specialrule{0.8pt}{2pt}{0pt}
\end{tabular}
}
\vspace{-3mm}
\end{table}

\begin{table*}[t!]
\centering
\caption{Quantitative comparison of \ac{DOR} baselines}
\label{tab:dor_performance_with_time}
\footnotesize
\setlength{\tabcolsep}{4.0pt}
\renewcommand{\arraystretch}{1.05}
\resizebox{\textwidth}{!}{
\begin{tabular}{l|cccc|cccc|cccc}
\specialrule{0.8pt}{0pt}{2pt}
\multirow{2}{*}{Method}
& \multicolumn{4}{c|}{\texttt{KITTI00}}
& \multicolumn{4}{c|}{\texttt{KITTI05}}
& \multicolumn{4}{c}{\texttt{AV2.0}} \\
\cmidrule(lr){2-5} \cmidrule(lr){6-9} \cmidrule(lr){10-13}
& SA(\%) $\uparrow$ & DA(\%) $\uparrow$ & AA(\%) $\uparrow$ & Time(s) $\downarrow$
& SA(\%) $\uparrow$ & DA(\%) $\uparrow$ & AA(\%) $\uparrow$ & Time(s) $\downarrow$
& SA(\%) $\uparrow$ & DA(\%) $\uparrow$ & AA(\%) $\uparrow$ & Time(s) $\downarrow$ \\
\specialrule{0.5pt}{2pt}{2pt}
Removert
& \textbf{99.44} & 41.53 & 64.26 & \underline{36.86}
& \textbf{99.42} & 22.28 & 47.06 & 182.94
& \textbf{98.74} & 34.10 & 58.02 & \underline{440.43} \\
ERASOR
& 66.70 & 98.54 & 81.07 & \textbf{5.57}
& 69.40 & \underline{99.06} & 82.92 & \textbf{12.42}
& 77.51 & \textbf{99.18} & \underline{87.68} & 508.66 \\
OctoMap
& 68.05 & \textbf{99.69} & 82.37 & 194.19
& 66.28 & \textbf{99.24} & 81.10 & 535.60
& 69.04 & \underline{97.50} & 82.04 & 5421.68 \\
OctoMap w/ GF
& 86.09 & \underline{98.84} & \underline{92.25} & 212.38
& 86.55 & 97.94 & \underline{91.88} & 570.60
& 77.88 & 86.18 & 81.92 & 5503.41 \\
\specialrule{0.5pt}{2pt}{2pt}
Elite++ (Ours)
& \underline{97.86} & 98.24 & \textbf{98.05} & 66.02
& \underline{98.39} & 94.47 & \textbf{96.41} & \underline{114.83}
& \underline{94.76} & 88.90 & \textbf{91.78} & \textbf{438.19} \\
\specialrule{0.8pt}{2pt}{0pt}
\end{tabular}
}
\vspace{-2mm}
\end{table*}

\subsection{Multi-session Experiments}

We evaluate multi-session static map maintenance on \texttt{ParkingLot01-02} and \texttt{KAIST01-04}.
As shown in \tabref{tab:registration_results}, ELite++ achieves RMSE and CD comparable to ELite, while reducing the processing time to about 11\% of the original runtime.
Considering that ELite requires an initial session-to-map candidate for alignment, these results show that ELite++ preserves alignment quality while automatically verifying local query submaps and fusing trusted window-level poses.

Furthermore, \figref{fig:multi} qualitatively shows the static maps generated on \texttt{ParkingLot01-02}.
ELite retains dynamic-object ghost artifacts, whereas ELite++ produces a cleaner static map by suppressing transient objects before integration.
These results indicate that ELite++ improves automation and update efficiency while maintaining alignment quality.
Detailed \ac{DOR} performance against existing baselines is evaluated in the next section.

\begin{figure}[!t]
    \centering
    \includegraphics[width=\linewidth]{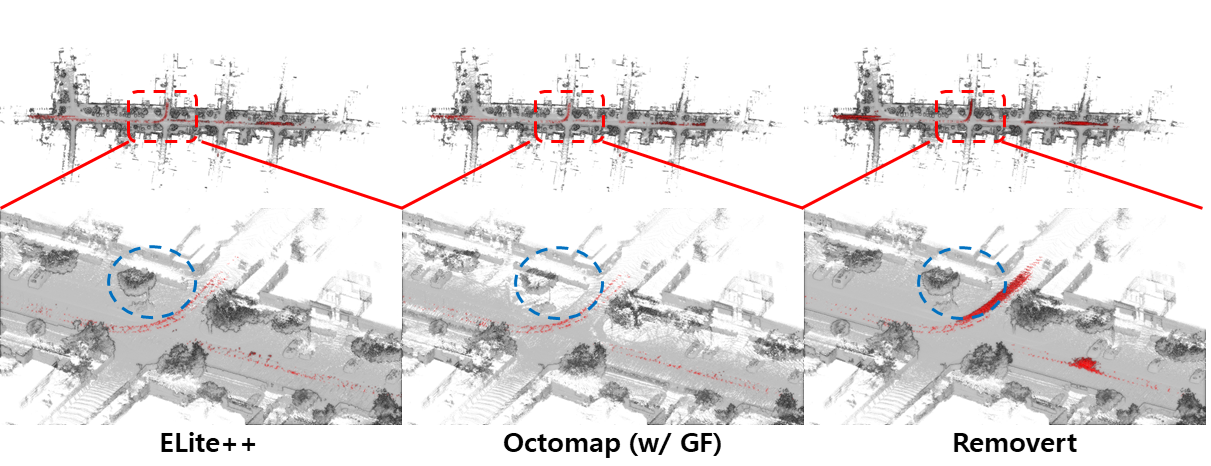}
    \caption{Qualitative \ac{DOR} comparison. Red points indicate remaining dynamic objects, while the blue region highlights over-removed static structures. ELite++ removes dynamic traces while better preserving static regions.}
    \label{fig:dynamic}
    \vspace{-4mm}
\end{figure}

\subsection{Dynamic Object Removal}

As shown in \tabref{tab:dor_performance_with_time}, ELite++ achieves the highest AA across all evaluated sequences.
Compared with OctoMap w/ GF, which is the strongest existing baseline on the KITTI sequences, our method consistently achieves better overall accuracy.
This shows that ELite++ provides a more balanced treatment of static and dynamic points.
Removert preserves static objects well but fails to sufficiently remove dynamic objects, whereas ERASOR and OctoMap show the opposite tendency by improving DA while sacrificing SA.
In contrast, ELite++ maintains both SA and DA at high levels, resulting in superior AA.

\figref{fig:dynamic} further illustrates this behavior qualitatively.
Removert leaves many unremoved dynamic objects, highlighted by red residual points, indicating insufficient dynamic filtering.
OctoMap w/ GF removes dynamic objects more aggressively and achieves a removal level comparable to ELite++, but it also removes static regions that should be preserved, as shown in the blue highlighted area.
By contrast, ELite++ suppresses dynamic traces while preserving nearby static structures, leading to a cleaner static map.

\begin{figure}[t!]
    \centering
    \includegraphics[width=.9\linewidth]{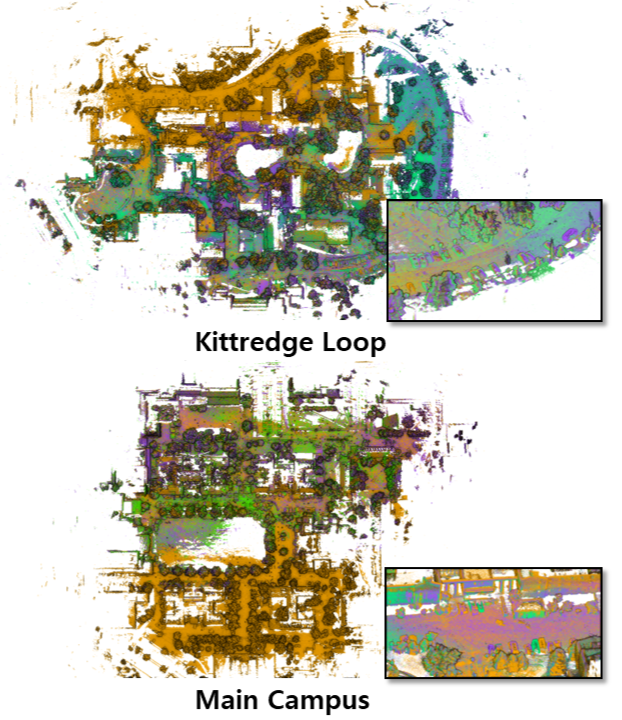}
        \caption{
    Qualitative multi-robot mapping results on the CU-Multi dataset.
    ELite++ aligns synchronized multi-run LiDAR sessions in \texttt{Main Campus} and \texttt{Kittredge Loop}.
    Different colors indicate different runs, and the zoomed regions show locally consistent structures in overlapping areas.
    }
    \label{fig:multi_robot}
\end{figure}

\subsection{Multi-Robot Experiments}
We evaluate ELite++ on the CU-Multi dataset to verify its applicability to multi-robot mapping scenarios.
The dataset provides synchronized multi-run LiDAR sequences in outdoor campus environments, including \texttt{Main Campus} and \texttt{Kittredge Loop}.
\figref{fig:multi_robot} shows qualitative mapping results.
Despite different trajectories and partial overlaps across runs, ELite++ aligns the sessions into a consistent map.
The zoomed regions show locally consistent structures in overlapping areas, indicating that the proposed window-based registration can provide reliable alignment for multi-robot static map maintenance.

\subsection{Ablation on Dynamic Object Removal}
\tabref{tab:dor_ablation} analyzes the role of each refinement in the proposed DOR module.
The voxel-wise baseline relies on local observation stability, which preserves many static points but often leaves dynamic objects only partially removed.
Static voxel recovery mitigates over-removal of persistent structures, while bottom-residue removal targets low-height dynamic parts that remain near the protected ground surface.
Cluster-level classification then resolves fragmented voxel decisions by enforcing spatial consistency over connected regions.
Together, these components balance static preservation and dynamic removal, leading to a cleaner static map with fewer residual artifacts.

\begin{table}[t]
\centering
\caption{Ablation results for DOR}
\label{tab:dor_ablation}
\footnotesize
\setlength{\tabcolsep}{5.0pt}
\renewcommand{\arraystretch}{1.05}
\resizebox{\columnwidth}{!}{
\begin{tabular}{ccc|ccc}
\specialrule{0.8pt}{0pt}{2pt}
Recovery & Bottom & Clust.
& SA(\%) $\uparrow$ & DA(\%) $\uparrow$ & AA(\%) $\uparrow$ \\
\specialrule{0.5pt}{2pt}{2pt}
$\times$ & $\times$ & $\times$
& 97.60 & 80.75 & 88.78 \\
$\checkmark$ & $\times$ & $\times$
& \textbf{98.70} & 80.74 & 89.27 \\
$\checkmark$ & $\checkmark$ & $\times$
& 98.67 & 82.25 & 90.09 \\
$\checkmark$ & $\checkmark$ & $\checkmark$
& 98.39 & \textbf{94.47} & \textbf{96.41} \\
\specialrule{0.8pt}{2pt}{0pt}
\end{tabular}
}
\end{table}

\subsection{Long-term Voxel Evidence}

We qualitatively analyze the long-term voxel evidence accumulated by the proposed Voxel Ratio Map.
Similar to the global ephemerality analysis in ELite~\cite{gil2025elite}, this experiment shows whether accumulated long-term evidence can indicate transient regions across multiple sessions.
Unlike ELite, which maintains point-wise global ephemerality, our method represents long-term occupancy stability using voxel-level hit and exposure statistics.

\begin{figure}[t]
    \centering
    \includegraphics[width=\linewidth]{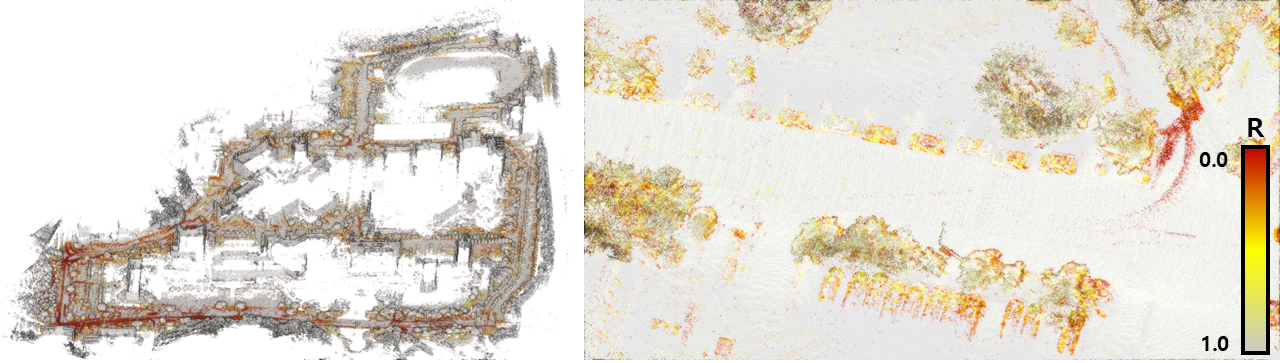}
    \caption{
Long-term voxel evidence from \texttt{KAIST01} and \texttt{KAIST04}.
Both the full map and zoomed view visualize accumulated voxel stability scores.
Low-stability regions correspond to intermittently observed objects, providing cues for dynamic-region filtering before final map generation.
}
    \label{fig:global_eph}
    \vspace{-4mm}
\end{figure}

\figref{fig:global_eph} shows the result constructed from \texttt{KAIST01} and \texttt{KAIST04}.
The visualization demonstrates that intermittently observed regions, such as parked or moving vehicles, obtain low stability scores, while persistent static structures remain highly stable.
This indicates that Voxel Ratio Maps can accumulate long-term static evidence and provide cues for dynamic-region filtering before final map generation.

\section{Conclusion}

This paper presented ELite++, an efficient lifelong LiDAR mapping framework for static map maintenance.
ELite++ combines automatic window-based registration, aligned-session \ac{DOR}, and compact Voxel Ratio Map integration to reduce update cost while preserving map quality.
Experiments on diverse LiDAR datasets show improved \ac{DOR} performance over existing baselines, maintained alignment quality, and runtime reduced to 11\% of the original pipeline.
These results demonstrate the practicality of ELite++ for continuous lifelong mapping in dynamic environments.

\bibliographystyle{IEEEtran}
\bibliography{refs}

@STRING{C-ECMR = {Proc. European Conf. on Mob. Robots}}

@STRING{C-ICRA = {Proc. {IEEE} Intl. Conf. on Robot. and Automat.}}

@STRING{C-IROS = {Proc. {IEEE}/{RSJ} Intl. Conf. on Intell. Robots and Sys.}}

@STRING{C-RSS = {Proc. Robot.: Science \& Sys. Conf.}}

@STRING{IEEE_J_IV = {{IEEE} Trans. Intell. Vehicles}}

@STRING{IEEE_J_RAL = {{IEEE} Robot. and Automat. Lett.}}

@STRING{IEEE_J_RO = {{IEEE} Trans. Robot.}}

@STRING{J-AR = {Autonomous Robots}}

@STRING{J-IJRR = {Intl. J. of Robot. Research}}

@STRING{J-JFR = {J. of Field Robot.}}

@article{cadena2016past,
  title={Past, present, and future of simultaneous localization and mapping: Toward the robust-perception age},
  author={Cadena, Cesar and Carlone, Luca and Carrillo, Henry and Latif, Yasir and Scaramuzza, Davide and Neira, Jos{\'e} and Reid, Ian and Leonard, John J},
  journal=IEEE_J_RO,
  volume={32},
  number={6},
  pages={1309--1332},
  year={2017}
}

@article{sousa2023systematic,
  title={A systematic literature review on long-term localization and mapping for mobile robots},
  author={Sousa, Ricardo B and Sobreira, Heber M and Moreira, Antonio Paulo},
  journal=J-JFR,
  volume={40},
  number={5},
  pages={1245--1322},
  year={2023}
}

@inproceedings{gil2025elite,
  title={Ephemerality Meets {LiDAR}-Based Lifelong Mapping},
  author={Gil, Hyeonjae and Lee, Dongjae and Kim, Giseop and Kim, Ayoung},
  booktitle=C-ICRA,
  year={2025}
}

@inproceedings{lim2021erasor,
  title={ERASOR: Egocentric ratio of pseudo occupancy-based dynamic object removal for static 3D point cloud map building},
  author={Lim, Hyungtae and Hwang, Sungwon and Myung, Hyun},
  booktitle=IEEE_J_RAL,
  volume={6},
  number={2},
  pages={2272--2279},
  year={2021}
}

@inproceedings{kim2021ltmapper,
  title={Lt-mapper: A modular framework for lidar-based lifelong mapping},
  author={Kim, Giseop and Kim, Ayoung},
  booktitle=C-ICRA,
  pages={7995--8002},
  year={2022}
}

@article{albin2025cumulti,
  title={CU-Multi: A Dataset for Multi-Robot Data Association},
  author={Albin, Doncey and Mena, Miles and Thomas, Annika and Biggie, Harel and Sun, Xuefei and Woods, Dusty and McGuire, Steve and Heckman, Christoffer},
  journal={arXiv preprint arXiv:2505.17576},
  year={2025}
}

@inproceedings{lim2025kissmatcher,
  title={Kiss-matcher: Fast and robust point cloud registration revisited},
  author={Lim, Hyungtae and Kim, Daebeom and Shin, Gunhee and Shi, Jingnan and Vizzo, Ignacio and Myung, Hyun and Park, Jaesik and Carlone, Luca},
  booktitle=C-ICRA,
  pages={11104--11111},
  year={2025}
}

@article{huang2021disco,
  title={Disco-slam: Distributed scan context-enabled multi-robot lidar slam with two-stage global-local graph optimization},
  author={Huang, Yewei and Shan, Tixiao and Chen, Fanfei and Englot, Brendan},
  journal=IEEE_J_RAL,
  volume={7},
  number={2},
  pages={1150--1157},
  year={2021}
}

@article{yin2023automerge,
  title={Automerge: A framework for map assembling and smoothing in city-scale environments},
  author={Yin, Peng and Zhao, Shiqi and Lai, Haowen and Ge, Ruohai and Zhang, Ji and Choset, Howie and Scherer, Sebastian},
  journal=IEEE_J_RO,
  volume={39},
  number={5},
  pages={3686--3704},
  year={2023}
}

@article{wei2024large,
  title={Large-scale multi-session point-cloud map merging},
  author={Wei, Hairuo and Li, Rundong and Cai, Yixi and Yuan, Chongjian and Ren, Yunfan and Zou, Zuhao and Wu, Huajie and Zheng, Chunran and Zhou, Shunbo and Xue, Kaiwen and others},
  journal=IEEE_J_RAL,
  volume={10},
  number={1},
  pages={88--95},
  year={2024}
}

@article{lim2025multi,
  title={Multi-Mapcher: Loop Closure Detection-Free Heterogeneous LiDAR Multi-Session SLAM Leveraging Outlier-Robust Registration for Autonomous Vehicles},
  author={Lim, Hyungtae and Kim, Daebeom and Myung, Hyun},
  journal=IEEE_J_IV,
  year={2025},
  publisher={IEEE}
}

@article{hornung2013octomap,
  title={OctoMap: An efficient probabilistic 3D mapping framework based on octrees},
  author={Hornung, Armin and Wurm, Kai M and Bennewitz, Maren and Stachniss, Cyrill and Burgard, Wolfram},
  journal=J-AR,
  volume={34},
  number={3},
  pages={189--206},
  year={2013}
}

@article{schauer2018peopleremover,
  title={The peopleremover—removing dynamic objects from 3-d point cloud data by traversing a voxel occupancy grid},
  author={Schauer, Johannes and N{\"u}chter, Andreas},
  journal=IEEE_J_RAL,
  volume={3},
  number={3},
  pages={1679--1686},
  year={2018}
}

@inproceedings{kim2020remove,
  title={Remove, then revert: Static point cloud map construction using multiresolution range images},
  author={Kim, Giseop and Kim, Ayoung},
  booktitle=C-IROS,
  pages={10758--10765},
  year={2020}
}

@inproceedings{lim2023erasor2,
  title={ERASOR2: Instance-Aware Robust 3D Mapping of the Static World in Dynamic Scenes.},
  author={Lim, Hyungtae and Nunes, Lucas and Mersch, Benedikt and Chen, Xieyuanli and Behley, Jens and Myung, Hyun and Stachniss, Cyrill},
  booktitle=C-RSS,
  year={2023}
}

@inproceedings{fan2022dynamicfilter,
  title={Dynamicfilter: an online dynamic objects removal framework for highly dynamic environments},
  author={Fan, Tingxiang and Shen, Bowen and Chen, Hua and Zhang, Wei and Pan, Jia},
  booktitle=C-ICRA,
  pages={7988--7994},
  year={2022}
}

@inproceedings{arora2021mapping,
  title={Mapping the static parts of dynamic scenes from 3D LiDAR point clouds exploiting ground segmentation},
  author={Arora, Mehul and Wiesmann, Louis and Chen, Xieyuanli and Stachniss, Cyrill},
  booktitle=C-ECMR,
  pages={1--6},
  year={2021},
  organization={IEEE}
}

@INPROCEEDINGS {Argoverse2,
  author = {Benjamin Wilson and William Qi and Tanmay Agarwal and John Lambert and Jagjeet Singh and Siddhesh Khandelwal and Bowen Pan and Ratnesh Kumar and Andrew Hartnett and Jhony Kaesemodel Pontes and Deva Ramanan and Peter Carr and James Hays},
  title = {Argoverse 2: Next Generation Datasets for Self-Driving Perception and Forecasting},
  booktitle = {Proceedings of the Neural Information Processing Systems Track on Datasets and Benchmarks},
  year = {2021}
}

@article{geiger2013vision,
  title={Vision meets robotics: The kitti dataset},
  author={Geiger, Andreas and Lenz, Philip and Stiller, Christoph and Urtasun, Raquel},
  journal=J-IJRR,
  volume={32},
  number={11},
  pages={1231--1237},
  year={2013}
}

@inproceedings{kim2020mulran,
  title={Mulran: Multimodal range dataset for urban place recognition},
  author={Kim, Giseop and Park, Yeong Sang and Cho, Younghun and Jeong, Jinyong and Kim, Ayoung},
  booktitle=C-ICRA,
  pages={6246--6253},
  year={2020}
}

@article{jung2024helipr,
  title={HeLiPR: Heterogeneous LiDAR dataset for inter-LiDAR place recognition under spatiotemporal variations},
  author={Jung, Minwoo and Yang, Wooseong and Lee, Dongjae and Gil, Hyeonjae and Kim, Giseop and Kim, Ayoung},
  journal=J-IJRR,
  volume={43},
  number={12},
  pages={1867--1883},
  year={2024}
}

\end{document}